\documentclass[letterpaper, 10 pt, conference]{ieeeconf}
\usepackage[english]{babel}
\IEEEoverridecommandlockouts                              

\usepackage{amsmath}
\usepackage{graphicx}
\usepackage{verbatim}
\usepackage[colorlinks=true, allcolors=blue]{hyperref}
\usepackage{multirow}
\usepackage{amsfonts}

\title{Generative Planning with Fast Collision Checks for High Speed Navigation}
\author{Craig Knuth, Cora A. Dimmig, Brian A. Bittner
\thanks{Johns Hopkins University Applied Physics Laboratory, Laurel, MD, USA.
        {\tt\small craig.knuth@jhuapl.edu}}
}
\begin{document}
\maketitle

\begin{abstract}
    Reasoning about large numbers of diverse plans to achieve high speed navigation in cluttered environments remains a challenge for robotic systems even in the case of perfect perceptual information. Often, this is tackled by methods that iteratively optimize around a prior seeded trajectory and consequently restrict to local optima. We present a novel planning method using normalizing flows (NFs) to encode expert-styled motion primitives. We also present an accelerated collision checking framework that enables rejecting samples from the prior distribution before running them through the NF model for rapid sampling of collision-free trajectories. The choice of an NF as the generator permits a flexible way to encode diverse multi-modal behavior distributions while maintaining a smooth relation to the input space which allows approximating collision checks on NF inputs rather than outputs. We show comparable performance to model predictive path integral control in random cluttered environments and improved exit rates in a cul-de-sac environment. We conclude by discussing our plans for future work to improve both safety and performance of our controller.
\end{abstract}

\section{Introduction}

As perception systems mature to handle high-speed navigation, motion planning at speed may become the dominating point of failure in the autonomy chain. We design a planner to safely complete navigation tasks in obstacle rich environments at competitive speeds using normalizing flows. Current planning frameworks do a poor job of considering quality multi-modal plans. This shortcoming is exacerbated in high speed scenarios, where higher obstacle density per planning horizon increases (in planar cases factorially) the number of homotopies\footnote{A homotopy, as defined in topology, is a grouping of paths between two points that are preserved under diffeomorphism.} to consider. We are proposing the first architecture we know of that directly incorporates the evasion tactics of experts and rapidly proposes hundreds of multimodal paths with low probability of collision.

Our approach applies generative modeling towards trajectory generation, permitting selection of maneuvers (styled by experience) spanning complex path modes (of arbitrary number) in real time. We utilize a normalizing flow (NF) architecture \cite{rezende2015variational}, which allows us to invertibly map samples from a prior distribution (in our work Gaussian) to a comprehensive distribution of observed motion primitives. 
We fit the NF to expert data and can sample expert-styled trajectories at rates amenable to real time motion planning (over 1kHz observed). 
Furthermore, this sampling is strongly biased to provide collision-free trajectories by masking out certain regions of the prior distribution. The specific masks are determined by an offline computation that roughly determines the samples in the prior distribution that would result in a collision for a given map.
Our method explicitly checks collision on only the highest performing motion primitives and consequently executes the best non-colliding trajectory.


We leverage two key insights in this work:
\begin{enumerate}
\item Generative networks contain the ability to sample a distribution of maneuvers that can be tailored to data without heuristics, fine tuning, or expert knowledge. The samples fit to expert data are generally aggressive and agile while naturally supporting the multi-modal inference suited to cluttered environments.
\item The invertibility and local smoothness of NFs allows for masking the prior distribution to strongly bias samples towards collision-free trajectories. This dramatically increases sample quality and requires far fewer costly collision checking operations.
\end{enumerate}
In summary, pairing a smooth generative model with a large cache of pre-computed collision checks results in a flexible approach to obtaining distributions of collision-free maneuvers that span the many planning modes presented in obstacle rich settings.

The outcome of our approach is a planner that excels at identifying a diverse set of multi-modal plans in real time. 
This method is juxtaposed with local planners that retain a notion of inertia such as Model Predictive Path Integral Control (MPPI) \cite{williams2016aggressive}) that are incapable of explicitly considering multiple homotopies simultaneously. We show that performance of both methods on random obstacle worlds are comparable, but that our generative planner excels in a trap-world environment (where bail-out maneuvers are challenging to identify for MPPI).

\section{Related Work}

High speed navigation in obstacle rich environments remains a significant challenge in the field of robotics. Classical motion planning and control architectures that are explorative in nature (e.g. Rapidly Exploring Random Tree (RRT) variants) are restricted in the quality of solution they can offer at competitive timescales. Recent edge processing enhancements have not improved this issue due to the sequential nature of the search. Classical exploitative solutions (e.g. Model Predictive Path Integral Control (MPPI) \cite{williams2016aggressive, pravitra2020ℒ}) is typically restricted to the homotopy of the seed trajectory due to the iterative nature of its optimizer. Similarly, Nonlinear Model Predictive Control (NMPC) and other gradient based methods are subject to local minima depending on the seed trajectory \cite{basescu2020direct, falanga2018pampc, li2021model}. Reinforcement Learning (RL) controllers \cite{hwangbo2017control} can be trained to encode expert maneuvers in real time, but have struggled to extend to agile maneuvers in cluttered environments. Additionally, RL controllers generally output a local trajectory distribution and cannot reason easily about multiple homotopies unless using an ensemble RL agent approach \cite{wiering2008ensemble}. 

Our method is similar to probabilistic motion primitives \cite{paraschos2018using,huang2019kernelized}, a sampling based alternative to the more traditionally used primitive library \cite{fod2002automated,schaal2005learning,pivtoraiko2011kinodynamic,bittner2021optimizing}. In our work we seek to build a planner for high speed obstacle dense navigation that is computationally feasible in real-time (unlike RRT), multi-modal and system agnostic (unlike NMPC), and performant with small datasets (unlike RL). 

\section{Problem Formulation}

We seek to produce a framework capable of high speed navigation through high density obstacle fields. In this work, we restrict our attention to planar examples but note that the method is directly extendable to full three-dimensional environments. We take the vehicle's initial pose $\xi_0 \in \Xi = SE(2)$, a running cost function $J$, an obstacle map $m \in M$, and a collision check function $c(x,m) \rightarrow \{0,1\}$ which determines if a state $\xi \in \Xi$ is colliding. We seek to compute and track a trajectory $\tau(t) \in \Xi,~t \in \mathbb{R}^+$ such that $\tau(0) = \xi_0$, $\tau$ minimizes the running cost function $J(\tau) \rightarrow \mathbb{R}$, and $\tau$ does not collide (i.e. $c(\tau(t),m)=0 \,\, \forall t$).

\section{Method}

We first collect an expert dataset of $N$ aggressive motion primitives $D = \{p_i\}_{i=1}^N$ and then train an NF $f$ to capture the distribution of the expert dataset. After generating the normalizing flow, we build a mask cache to roughly identify colliding samples in input space. Online we use the mask cache to only run samples from the prior distribution through the NF if the samples are biased towards inhabiting free space. 
We select the best non-colliding primitive at each planning iteration (only explicitly calculating collision for the best primitives), and track that primitive with a low level controller until the next planning iteration.

In practice, large input and output dimensions of NFs can lead to challenges in training, runtime performance, and masking input regions for collision avoidance. Consequently, it is important to parameterize the expert primitives with as few variables as possible while also still capturing the variability inherent in the dataset. In this work, we fix the expert primitives $p_i$ to be equal length in time and parameterize the primitives as three consecutive equal-length arcs each with their own curvature. This yields a four dimensional parameterization $\theta = \{\alpha, \kappa_1, \kappa_2, \kappa_3\}$ where $\alpha$ is the length of the primitive and $\kappa_i$ are the curvatures of each equal-length arc. Let $p(\theta)$ be the reconstruction of the primitive in state space.

The resulting NF maps an input $z \in Z = \mathbb{R}^d$ to a primitive parameterization $\theta \in \Theta = \mathbb{R}^d$.
\begin{equation}
    f : z \rightarrow \theta, \quad z \sim N_d(0,1)
\end{equation}

Here our prior distribution is the standard $d$ dimensional Gaussian, $N_d(0,1)$. $f$ is trained with stochastic gradient descent to minimize the loss $L$ over sample batches $B$ taken from our prior. Below $KL$ is the reverse Kullback-Leibler divergence.
\begin{equation}
    L(B) = KL(f(B), D)
\end{equation}

Our mask cache relies on discretizing both the input space $Z$ and obstacle maps $M$.
Additionally, obstacle maps are composed from so-called atomic obstacle maps $m_i$. An atomic obstacle map contains a single obstacle (in this work, a circular obstacle with fixed radius) that is aligned to a grid point in the body frame of the robot. Online, an obstacle map $m$ can be decomposed into a set of atomic maps $m_i$. The set of atomic maps can either over- or under-cover the full map. In this work, we chose to minimally over-cover the full map such that $m \subset \bigcup_i m_i$ which produces fewer colliding trajectories at the expense of potentially excluding some non-colliding trajectories.

Our mask cache is built with the following structure. We discretize the input space into $K$ bins in each dimension which correspond to our masks denoted $\{\mu_j\}_{j=1}^{K \cdot d}$. The bin extents are chosen so that each bin is equally likely to contain a sample from the prior. Then, the centroid of each mask $\mu_j$, denoted $z_j$, is pushed through the normalizing flow to produce a primitive $p(f(z_j))$. For each atomic map $m_i$, we determine all the masks that produce a colliding primitive from its centroid, i.e. all $\mu_j$ such that there exists $\xi \in p(f(z_j))$ where $c(\xi, m_i)=1$.

Online, we decompose a map $m$ into the atomic maps. Then we use the mask cache to look up all masks that have a colliding centroid with at least one atomic map. At each planning iteration, we reject any samples that lie in one of these masks. Note this rejection happens in the input space before running the normalizing flow.

\begin{figure}
    \centering
    \includegraphics[width=0.48\textwidth]{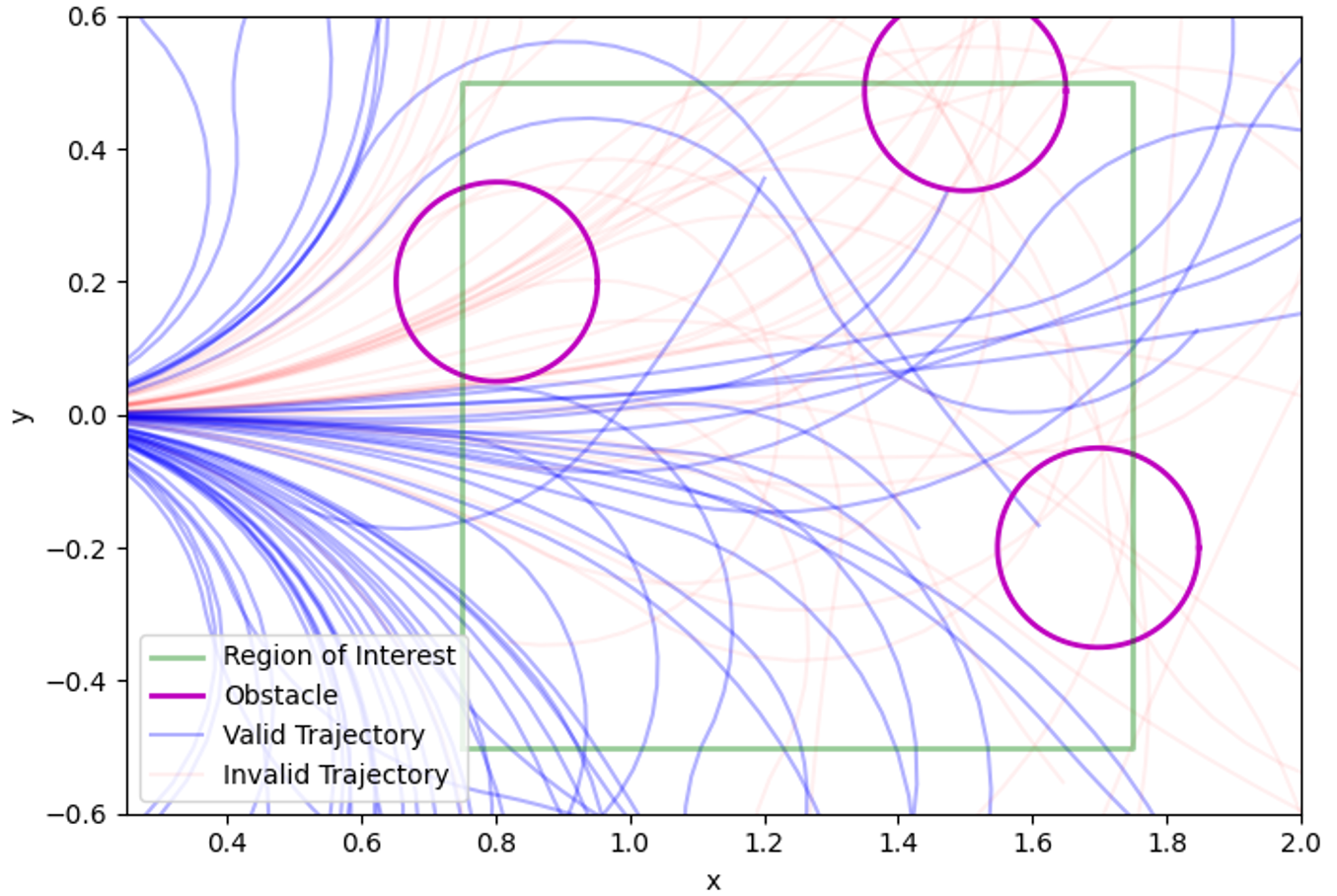}
    \caption{Samples for a single planning iteration. The green square identifies the region of interest where masking is performed. Red trajectories are masked, blue are not. Notice that one blue primitive goes through the lower obstacle; this primitive would be checked for collision if it was low cost. Also notice the samples are diverse across the homotopies of the environment.}
    \label{fig:plan}
\end{figure}

Using the centroid to determine collision for all $z$ in a mask $\mu_j$ provides only a coarse assessment of collision. With this approach, we are not excluding all colliding primitives nor including all non-colliding primitives. To ensure safety of the executed primitive, we compute the cost for all sampled primitives and check collision for only the lowest cost primitives until we determine the lowest cost non-colliding sampled primitive. This approach allows for far fewer samples to be pushed through the NF and collision checks (as only the lowest cost primitives are checked). We illustrate a single planning iteration of our controller in Figure \ref{fig:plan}.

\begin{figure}
    \centering
    \includegraphics[width=0.5\textwidth]{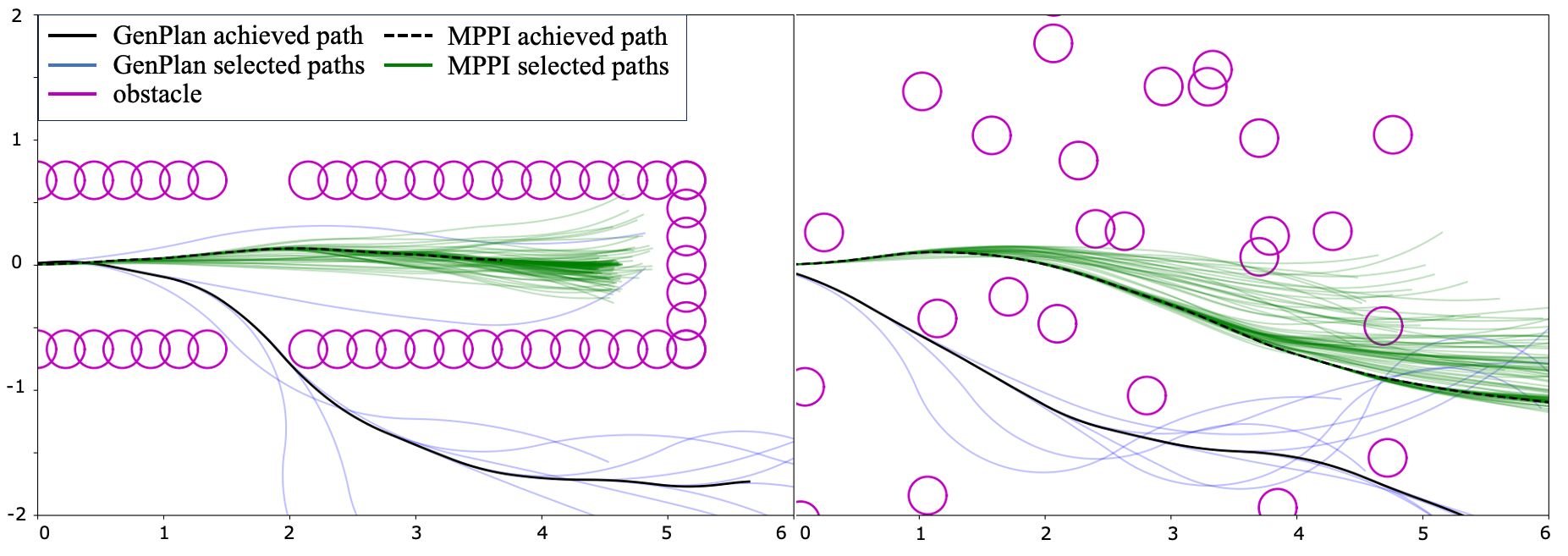}
    \caption{Example rollouts in cul-de-sac (left) and random (right) worlds. Solid black is the rollout by GenPlan with selected plans at each planning iteration shown in blue. Dashed black is the rollout by MPPI with selected plans at each planning iteration shown in green. GenPlan is capable of exiting the cul-de-sac environment at much higher rates than MPPI.}
    \label{fig:rollouts}
\end{figure}

\section{Experimental Setup}

To build the expert dataset, we teleoperated a lightly modified MITRacecar \cite{cain2017high} platform and collected motion capture data via OptiTrak for about 30 minutes. 
This data is organized into a collection of 836 primitives by slicing the full motion capture trajectory into two second subtrajectories. We then fit parameters $\theta \in \Theta$ to each subtrajectory using least squared error on position. We fit a neural spline model \cite{durkan2019neural} with two layers of two blocks and 16 hidden channels. This two-layer neural spline architecture was selected after a performance comparison against planar flow \cite{rezende2015variational}, NICE \cite{dinh2014nice}, real NVP \cite{dinh2016density}, and residual flow \cite{chen2019residual} architectures considering both final loss and computational efficiency.

The input space $Z$ is discretized into 40 bins per dimension each with equal sample probability. We only consider obstacles in an area 75 to 175 cm in front of the vehicle and within 50 cm laterally. This area is chosen to reduce the size of the mask cache. The space of atomic maps is discretized into 40 bins along each of the map's dimensions in x and y (yielding obstacles every 2.5 cm in x and y). Each atomic map has a singular circular obstacle with a radius of 15 cm. We sample this generative model 512 times running at 5 Hz. The lowest cost non-colliding primitive is tracked by a PID controller at 100 Hz until the next planning iteration.

We evaluate our method in two 2.5 second simulated scenarios (shown in Figure \ref{fig:rollouts}) compared to the baseline MPPI (see appendix \ref{app:mppi}) as presented in \cite{nagabandi2020deep}. The vehicle dynamics are governed by the bicycle model (see appendix \ref{app:bike}). In both scenarios, the vehicle starts at $x_0 = [x, y] = [-0.5, 0]$ with initial velocity 2.5 m/s and heading in the positive $x$ direction. The cost is given by $J(x) = -x$, i.e. the goal is to drive in the positive $x$ direction. Our first scenario is a random obstacle world where 50 circular obstacles with radius 15 cm are placed uniformly at random between $[0, 5]$ in $x$ and $[-3, 3]$ in $y$. The second scenario is a classic cul-de-sac environment.

\section{Results}

In the random obstacle world, our controller GenPlan performs comparably to MPPI in both positive $x$ progress and collision rates, see Table \ref{tab:random}. Terminal $x$ and average velocity metrics are calculated only on non-colliding trials.
In the cul-de-sac world, GenPlan exits at a much higher rate than MPPI, 50\% compared to 11\%. See Table \ref{tab:trap}. We attribute this to GenPlan's diverse sampling capability and lack of history. Since MPPI by design is an iterative optimizer, it struggles to quickly change plans or consider multiple planning homotopies at the same time. 
We note that MPPI does not enjoy higher exit rates with greater sampling noise. We performed additional experiments on the cul-de-sac map with a variety of sampling noises; those results are given in appendix \ref{app:sweep}.

\begin{table}[h]
\centering
\caption{Results over 100 trials in random map. Mean $\pm$ std dev.}
\label{tab:random}
\vspace{-12pt}
\begin{center}
\renewcommand{\arraystretch}{1.3}
\begin{tabular}{ c || c c c }
 Controller & Collision \% & Terminal $x$ [m] & Avg Vel [m/s] \\ 
 \hline\hline
  GenPlan & 6\% & 6.33 $\pm$ 0.30 & 2.63 $\pm$ 0.06 \\  
 MPPI & 4\% & 6.59 $\pm$ 0.80 & 2.77 $\pm$ 0.21
\end{tabular}
\end{center}
\vspace{-12pt}
\end{table}

\begin{table}[h]
\centering
\caption{Results over 100 trials in cul-de-sac map. Mean $\pm$ std dev.}
\label{tab:trap}
\vspace{-12pt}
\begin{center}
\renewcommand{\arraystretch}{1.3}
\begin{tabular}{ c || c c c c }
 \multirow{2}{4.2em}{\centering Controller} &
 \multirow{2}{3.8em}{\centering Exit rate \%} & 
 \multirow{2}{3.8em}{\centering Collision \%} & 
 \multirow{2}{5em}{\centering Terminal $x$ [m]} & 
 \multirow{2}{4em}{\centering Avg Vel [m/s]} \\ \\
 \hline\hline
 GenPlan & 50\% & 0\% & 5.14 $\pm$ 1.00 & 2.17 $\pm$ 0.48 \\  
 MPPI & 11\% & 0\% & 4.42 $\pm$ 0.56 & 1.80 $\pm$ 0.30
\end{tabular}
\end{center}
\vspace{-12pt}
\end{table}

\section{Conclusions and Future Work}

In this abstract, we present a planning framework using NFs to sample expert-styled motion primitives. The primitive sampling is accelerated by masking poor samples prior to running the model through a mask cache which consequently strongly biases the sampling to produce diverse and non-colliding trajectories. We present a comparison to MPPI and show comparable performance in random cluttered environments but much higher exit rates in a cul-de-sac environment.

While the results are encouraging, we seek to improve performance of the GenPlan controller. We will test our results in hardware where we anticipate improved performance of GenPlan relative to MPPI as MPPI will no longer enjoy perfect information with regards to the vehicle dynamics.

Futhermore, we will improve the safety margin of the GenPlan controller by certifying tracking performance using techniques established in recent work \cite{knuth2021planning, knuth2023statistical, chou2021model} that applies Extreme Value Theory (EVT) to certifying control with learned dynamics. In our case, we will collect data tracking primitives provided by the NF, and then fit a learned tracking error profile on the primitive. This tracking error will be certified with EVT and be used to assert obstacle avoidance even in the case of tracking error.


\bibliographystyle{IEEEtran}
\bibliography{citations}

\appendix

\subsection{Bicycle Dynamics} \label{app:bike}

The bicycle dynamics used in experiments are defined as follows. The state is defined as position in $x$, position in $y$, linear velocity $v$, heading $\phi$, and steering angle $\psi$ with inputs acceleration $u_a$ and steering angle velocity $u_{\dot{\psi}}$. The dynamics are given as: $\dot{x} = v \cos(\phi)$, $\dot{y} = v \sin(\phi)$, $\dot{v} = u_a$, $\dot{\phi} = v \tan(\psi)$, and $\dot{\psi} = u_{\dot{\psi}}$.


\subsection{MPPI Implementation} \label{app:mppi}

MPPI uses the bicycle dynamics to rollout trajectories and runs at 50 Hz. At each iteration it samples 1024 trajectories, and plans 2 seconds ahead. The initial control trajectory given to MPPI is simply all zeros which corresponds to driving forward at the same velocity. The sampling variance is tuned to $2$ in acceleration and $16\pi/5$ in steering angle velocity. The sampling noise correlation factor is set to $\beta=0.25$, and the exponential reward weighting factor is set to $\gamma=1$.


\subsection{MPPI Performance on Cul-de-Sac} \label{app:sweep}

We additionally evaluate MPPI's performance in the cul-de-sac environment according to a variety of sampling noises. Specifically, we investigate if MPPI can achieve greater exit rates in the bug trap world if it has higher sampling noise, potentially at the cost of performance in the random obstacle world. The results reported in the main section are the best results MPPI achieved across the tested sampling noises. Results on additional sampling noises are given in Table~\ref{tab:mppi_sigma_trap} where the $\sigma$ values correspond to the sample variance for $u_a$ and $u_{\dot{\psi}}$ respectively. We do not report values for the lowest $\sigma$ since all trials resulted in collision. 
We conclude that MPPI is fundamentally limited by its initial seed in the cul-de-sac environment and could not achieve exit rates similar to GenPlan even for optimally tuned sampling noises.

\begin{table}[h]
\centering
\caption{Results over 100 trials for different MPPI sigma values in cul-de-sac map. Mean $\pm$ std dev.}
\label{tab:mppi_sigma_trap}
\vspace*{-2mm}
\begin{center}
\renewcommand{\arraystretch}{1.3}
\begin{tabular}{ c || c c c c }
 \multirow{2}{4em}{\centering$\sigma$} & 
 \multirow{2}{3.8em}{\centering Exit rate \%} & 
 \multirow{2}{3.8em}{\centering Collision \%} & 
 \multirow{2}{5em}{\centering Terminal $x$ [m]} & 
 \multirow{2}{4em}{\centering Avg Vel [m/s]} \\ \\
 \hline\hline
 (0.1, 0.16$\pi$)& 0\%& 100\%& - & - \\
 (0.5, 0.8$\pi$)& 0\%& 0\%& 4.48 $\pm$ 0.03&1.80 $\pm$ 0.01\\
 (1, 1.6$\pi$)& 0\%& 0\%& 4.39 $\pm$ 0.02&1.76 $\pm$ 0.01\\
 (1.5, 2.4$\pi$)& 4\%& 0\%& 4.39 $\pm$ 0.34&1.77 $\pm$ 0.18\\
 (2, 3.2$\pi$)& 11\%& 0\%& 4.42 $\pm$ 0.56&1.80 $\pm$ 0.30\\
 (2.5, 4$\pi$)& 3\%& 0\%& 4.16 $\pm$ 0.32&1.67 $\pm$ 0.17\\
 (3, 4.8$\pi$)& 0\%& 0\%& 3.92 $\pm$ 0.10&1.57 $\pm$ 0.04\\
 (3.5, 5.6$\pi$)& 1\%& 0\%& 3.73 $\pm$ 0.12&1.51 $\pm$ 0.12\\
 (4, 6.4$\pi$)& 1\%& 0\%& 3.48 $\pm$ 0.25&1.41 $\pm$ 0.16\\
 (4.5, 7.2$\pi$)& 0\%& 0\%& 3.21 $\pm$ 0.21&1.29 $\pm$ 0.08\\
 (5, 8$\pi$)& 1\%& 0\%& 2.93 $\pm$ 0.23&1.20 $\pm$ 0.08\\
\end{tabular}
\end{center}
\end{table}

\end{document}